\documentclass[letterpaper, 10 pt, conference]{ieeeconf}  

\IEEEoverridecommandlockouts                              

\usepackage{graphics} 
\usepackage{times} 
\usepackage{amsmath} 
\usepackage{amssymb}  
\usepackage{bm}

\usepackage{subcaption}
\usepackage{graphicx}
\usepackage{xcolor}
\usepackage{booktabs}
\usepackage{xspace}
\usepackage[ruled,linesnumbered]{algorithm2e}
\usepackage{url}

\newcommand{\name}{PMR}

\newcommand{\hide}[1]{}

\title{\LARGE \bf
Learning Design and Construction with Varying-Sized Materials\\via Prioritized Memory Resets
}

\author{Yunfei Li$^{1}$, Tao Kong$^{2}$, Lei Li$^{3}$ and Yi Wu$^{1,4}$
\thanks{$^{1}$Institute for Interdisciplinary Information Sciences, Tsinghua University, Beijing, China.
        {\tt\small \{liyf20@mails,jxwuyi@mail\}.tsinghua.edu.cn}}%
\thanks{$^{2}$ByteDance AI Lab, Beijing, China.  {\tt\small kongtao@bytedance.com}
}%
\thanks{$^{3}$University of California Santa Barbara. {\tt\small lilei@cs.ucsb.edu}
}%
\thanks{$^{4}$Shanghai Qi Zhi Institute, Shanghai, China}
}

\let\oldtwocolumn\twocolumn
\renewcommand\twocolumn[1][]{%
    \oldtwocolumn[{#1}{
    \begin{center}
          \includegraphics[width=\textwidth]{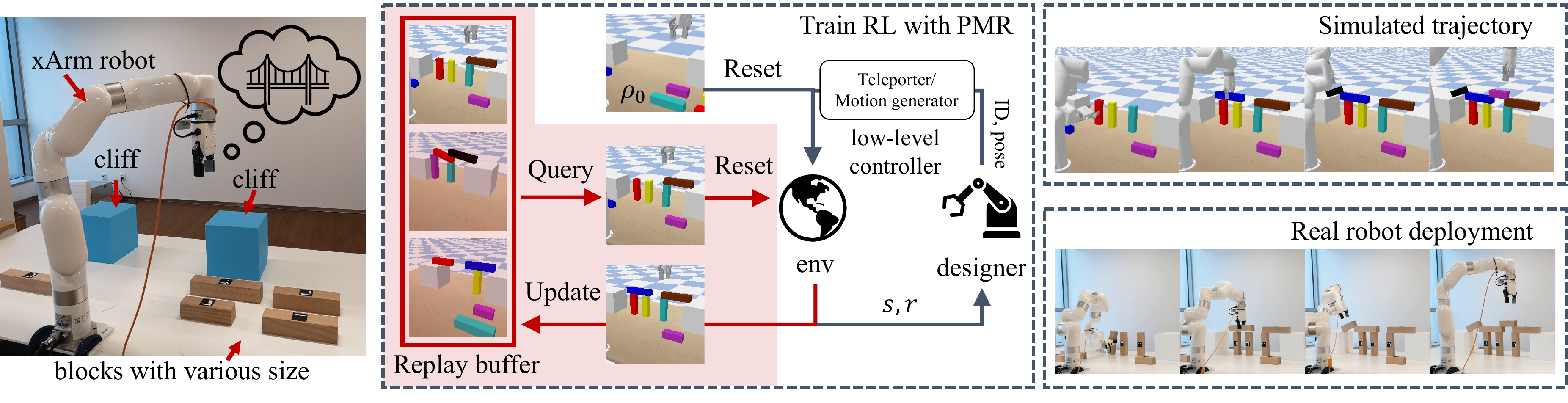}
          \captionof{figure}{Bridge design and construction with a collection of varying-sized objects. \textit{Left}: Task setting. \textit{Middle}: Overview of training approach. An RL designer is trained with \underline{P}rioritized \underline{M}emory \underline{R}esets (red shadowed area). The predicted instruction from designer is executed with a low-level controller.
          \textit{Right}: Evaluation in simulation and on a real robot. 
          }
          \label{fig:teaser}
    \end{center}
    }]
}

\begin{document}

\maketitle
\thispagestyle{empty}
\pagestyle{empty}

\begin{abstract}
\hide{
We consider a particularly challenging robotic manipulation task, \emph{bridge design and construction}, where the robot is provided with a set of building blocks in various sizes and needs to pick specific blocks to build a bridge over the cliffs without a given blueprint. This is a long-horizon task with sparse reward, and, due to the diverse block sizes, the exploration issue is particularly severe for learning. To tackle this challenge, we adopt a hierarchical solution, consisting of a high-level reinforcement-learning-based designer to propose object-centric building instructions and a low-level motion-planning-based action generator to manipulate a particular block. For effective reinforcement learning, we propose a novel learning paradigm, \emph{Prioritized Memory Resets} (\name), which significantly improves exploration. {\name} adaptively resets the environment state to those most critical intermediate configurations from the replay buffer so that the robot can re-start training on a half-done architecture instead of from scratch. Furthermore, we also combine {\name} with auxiliary training objectives and fine-tune the designer with the locomotion generator, which eventually yields an intelligent robot system that can effectively handle the bridge construction task with various block sizes in the real world. Demos can be found at \url{https://sites.google.com/view/bridge-pmr}.
}
Can a robot autonomously learn to design and construct a bridge from varying-sized blocks without a blueprint? It is a challenging task with long horizon and sparse reward -- the robot has to figure out physically stable design schemes and feasible actions to manipulate and transport blocks. Due to diverse block sizes, the state space and action trajectories are vast to explore. In this paper, we propose a hierarchical approach for this problem. It consists of a reinforcement-learning designer to propose high-level building instructions and a motion-planning-based action generator to manipulate blocks at the low level. For high-level learning, we develop a novel technique, \emph{prioritized memory resetting} ({\name}) to improve exploration. {\name} adaptively resets the state to those most critical configurations from a replay buffer so that the robot can resume training on partial architectures instead of from scratch. Furthermore, we augment {\name} with auxiliary training objectives and fine-tune the designer with the locomotion generator. Our experiments in simulation and on a real deployed robotic system demonstrate that it is able to effectively construct bridges with blocks of varying sizes at a high success rate. Demos can be found at \url{https://sites.google.com/view/bridge-pmr}.
\end{abstract}

\IEEEpeerreviewmaketitle

\section{Introduction}
Reinforcement learning (RL) has been an increasingly promising paradigm for solving complex robotic manipulation tasks~\cite{gu2017deep,rajeswaran18dexterous,akkaya2019solving,yu2020meta}, such as grasping~\cite{kalashnikov2018qt}, stacking~\cite{nair18overcoming,li2020towards}, object rearrangement~\cite{openai2021asymmetric}, mobile manipulation~\cite{li2019hrl4in,xiarelmogen} and folding towels~\cite{balaguer2011combining}. For many tasks that are overly complicated for classical control methods, RL serves as a general solution to enable a learning robot to automatically solve them assuming minimal domain knowledge.

In this paper, we tackle a challenging task, \emph{bridge design and construction with varying-sized objects} (Fig.~\ref{fig:teaser}), where a collection of \emph{varying-sized} building blocks are given while the robot needs to select necessary blocks to construct a stable bridge connecting two distant cliffs. 
In contrast with many manipulation tasks in the existing literature, where the target object configuration is often known in advance~\cite{luo19assembly,zhu2020hierarchical},
in this task, neither the target bridge architecture nor the construction instructions are given. The robot needs to adaptively manipulate blocks with appropriate sizes in the right order to finish construction based on the given materials and the cliff width. Moreover, the success signal can be only obtained after a valid bridge is completely built. 
Hence, this problem is particularly difficult for its \emph{long horizon} and \emph{sparse reward}, and requires non-trivial  \emph{exploration} due to its varying block sizes and cliff width. 
A similar bridge design and construction variant is studied in~\cite{li2021learning}, but is less challenging since it only considers blocks of identical sizes and utilizes hand-crafted intermediate dense reward.


To tackle this challenge, we adopt a hierarchical solution consisting of an RL-based high-level designer and a planning-based low-level action generator. The high-level designer is trained by RL over object-centric states to decompose the long-horizon task into a sequence of single-block pick-and-place sub-tasks. The low-level policy simply executes the pick-and-place instructions by motion planning and produces a collision-free sequence of robot actions. 
This is conceptually similar to the ReLMoGen framework~\cite{xiarelmogen}. ReLMoGen focuses on the mobile robot setting and produces state-based sub-goals, such as spatial positions and arm states, while we consider object configurations.
In addition, ReLMoGen trains the high-level planner with a locomotion generator throughout the entire training process, which, however, is computationally expensive and results in poor exploration in our task. We propose to first train the high-level designer solely by teleporting the selected block to the target position and then fine-tune the designer with the action generator, which achieves a substantially higher success rate.



To effectively overcome the exploration challenge when training the high-level designer, we propose a novel training paradigm, \emph{Prioritized Memory Resets} (\name). 
{\name} adaptively resets the RL environment to a past state selected from the replay buffer, so that the robot can start from an intermediate half-done architecture instead of always restarting from scratch in each training episode. The insight of {\name} is that in this sparse-reward hard-exploration problem, some close-to-success configurations may be hardly re-achieved from scratch by the training policy. Therefore, we can directly reset the environment to those states that may lead to the biggest learning advancements.
{\name} is conceptually similar to automatic curriculum learning methods in goal-conditioned RL~\cite{florensa2017reverse,florensa2018automatic}, which trains the agent with adaptive \emph{target goals} for fast policy improvements. By contrast, {\name} resets the \emph{initial state} of a training episode.
In addition to {\name}, we also adopt an auxiliary self-supervised objective for better representation learning, which further accelerates learning.


Experiment result shows that our RL-based bi-level solution achieves a success rate of 71.8\% in simulation for constructing a bridge with 7 random-sized blocks and discovers interesting bridge architectures while the standard RL method fails completely. Ablation studies also demonstrate that each of our algorithmic components, including {\name}, self-supervised learning, and locomotion fine-tuning, is critical to the overall performance. We also validate our method on a real-world robot arm. 


Our contributions are summarized as follows:
\begin{enumerate}
    \item We develop an effective hierarchical RL framework to tackle the long-horizon sparse-reward varying-sized bridge construction problem.
    \item We propose a novel technique, \emph{Prioritized Memory Reset} (\name), to tackle the exploration challenge in long-horizon sparse-reward robot learning problems.
    \item We show that self-supervised learning and locomotion fine-tuning substantially improves RL training in complex manipulation tasks.
    \item We develop an effective RL-based robot system that can successfully handle the challenging 
    varying-sized bridge design and construction task in the real world.
\end{enumerate}


\section{Related Work}
Robot construction and manipulation tasks~\cite{knepper2013ikeabot,nagele20lego,zakka2020form2fit,haarnoja2018composable,li2020towards,zhu2020hierarchical,batra20rearrangement} serve as a popular testbed for developing intelligent manipulators with long-term autonomy.
Most of the construction tasks assume a known target state in a priori, i.e., the desired configuration designed by a human expert is provided to the robot.
We focus on a bridge design and construction task with no prior knowledge of the precise target state. So, the robot has to both design the bridge architecture and construct the bridge via a sequence of feasible control actions.
There are also works focusing on generating structure designs under particular constraints without considering  construction~\cite{ritchie2015generating,ritchie2015controlling}. 

Hierarchical frameworks are commonly adopted in complex long-horizon manipulation tasks. 
Hierarchical reinforcement learning (HRL) typically learns a bi-level policy, with the high-level policy generating sub-goals for the low-level policy to execute~\cite{kulkarni2016hierarchical,bacon17option}. 
The two policies can be optimized jointly~\cite{nachum18data-efficient,nachum19near,bagaria2020option} or separately~\cite{heess16learning}. 
ReLMoGen~\cite{xiarelmogen} tackles mobile manipulation and interactive navigation tasks with a combination of a learnable high-level policy and a fixed low-level motion generator.  
Our framework is conceptually similar to ReLMoGen, with a high-level RL designer integrated with a classical motion generator while our high-level policy considers object configurations instead of spatial positions or robot states. Moreover, we only use locomotion to fine-tune the high-level policy for better designer exploration while ReLMoGen leverages both parts throughout training.

The proposed prioritized memory reset technique adaptively proposes critical intermediate states for the agent to restart from, 
which is related to automatic curriculum learning methods that propose tasks with moderate difficulty~\cite{zhang2020automatic,florensa2017reverse,florensa2018automatic,sukhbaatar2018intrinsic}. 
However, these methods mainly work for goal-conditioned problems or a fixed set of tasks by generating goals, while we directly reset the environment to previously visited states. 
Also, the curriculum learning methods in goal-conditioned RL typically assume a known goal space, while {\name} does not need to explicitly know the space of state configurations.
{\name} enhances exploration by directly teleporting the agent to previous states without changing the reward function, which is different from classical intrinsic-reward-based exploration~\cite{singh2005intrinsically,burda19rnd,pathak18largescale}.
{\name} is most related to Go-Explore~\cite{ecoffet2021first}, which also teleports the agent to promising past states. However, Go-Explore uses a count-based metric as its state selection criterion, which exhaustively explores the entire state space. This is infeasible in complex construction tasks where exponentially many failure/unstable states exist and only a few architectures are crucial for success. Hence, {\name} adopts a value-error-based criterion, which gradually learns to only focus on stable states as training proceeds. 
We remark that some model-based RL methods~\cite{schrittwieser2020mastering} also restore a visited state to perform monte-carlo tree search. However, these methods require an accurate forward model while {\name} is model-free. 
Finally, we use inverse dynamics prediction as an auxiliary task for better representation learning. This self-supervised objective has been also applied in other problems including exploration~\cite{pathak17curiosity} and meta-learning~\cite{hansen21self}.


\section{Task Setup}
In our bridge design and construction task, there are $N$ building blocks on the table and two ``cliffs'' at a random distance from each other. 
All the building blocks are of width and height 5cm while their lengths are sampled from the following three categories: standard length $L$ which is equal to the cliff height; shorter length uniformly sampled from $0.5L$ to $0.9L$; longer length uniformly sampled from $1.1L$ to $1.25L$. A 7 DoF xArm robot is mounted on the side of the table to manipulate the building blocks. The robot aims to design and construct a bridge using the building blocks that can connect the cliffs. The simulated environment is built with PyBullet~\cite{coumans2020}.
We tackle this long-horizon manipulation task with a hierarchy of a high-level RL-based designer that sequentially instructs one object to a new pose at each time, and a low-level controller that generates collision-free robot motions to accomplish the high-level instructions.

\subsection{Problem Formulation}
The low-level control problem is a standard motion-planning task while the high-level bridge design problem is formulated as a Markov decision process defined as follows:

\textbf{Observation:} In each step, the agent observes the positions, orientations, velocities, and sizes of all the building blocks and cliffs. 

\textbf{Action:} The agent can instruct one building block to a new pose within the vertical plane that goes through the centers of cliffs. Each action is a vector of 4 elements $(\textrm{ID}, y, z, \textrm{angle})$, where $\textrm{ID}$ denotes which block to move, $y$ and $z$ specify the target position of the block's center of mass, $\textrm{angle}$ denotes the 1-D rotation within the plane.

\textbf{Reward:} For each step, the agent can get a $0.1$ reward only if a bridge is successfully built; otherwise it receives no reward. 
We cast multiple rays downwards onto the structure inside the valley to detect its height.
If the height of all the detected points is greater than the cliff height plus the block thickness, we consider the structure successful.

\textbf{Horizon: } Each episode lasts a fixed length of 30 steps.  

\textbf{Initial state distribution $\rho_0$:} When the environment resets, the distance between the cliffs is sampled from $[0.75L, 3.75L]$. We then sample a set of building blocks consisting of $\lfloor N/2 \rfloor$ standard blocks, $\lceil N/4 \rceil$ long blocks, and $(\lceil N/2 \rceil - \lceil N/4 \rceil)$ short blocks. All the building blocks are aligned on the table outside the valley. 

\subsection{Policy Architecture}
The high-level RL designer is instantiated as an actor $\pi_{\theta}$ and a critic $V_{\phi}$ with a shared transformer-based~\cite{Vaswani17attention} encoder $\psi$. $\psi$ stacks 3 self-attention blocks to extract object features $h_t^{[N+2]}$ from $s_t^{[N+2]}$, where the superscript $[i]$ denotes a sequence from $1$ to $i$ and the subscript $t$ denotes the time step. Then we feed $h_t$ into $\pi_{\theta}$ and $V_{\phi}$ to get action and value prediction. 
$\pi_{\theta}$ models a joint action distribution $p(\textrm{ID}, y, z, \textrm{angle}) = p(\textrm{ID}) p(y|\textrm{ID}) p(z|\textrm{ID}) p(\textrm{angle}|\textrm{ID})$. 
$p(\textrm{ID})$ is represented as a categorical distribution of $(N + 1)$ classes. The first class represents a special ``no-op'' action, which means that the agent will not move any object in this step. It is useful in the steps when the bridge is successfully built. The remaining $N$ classes correspond to moving different building blocks. The raw distribution (logit) of ``ID'' is computed as $\textrm{logit}(\textrm{ID}=0) = g_{\textrm{no-op}}(\sum_{i=1}^{N+2} h_t^i / N)$, $\textrm{logit}(\textrm{ID}=i) = g_{\textrm{ID}}(h_t^i), i=1, 2, \cdots, N$, where $g_{\textrm{no-op}}$ and $g_{\textrm{ID}}$ are linear layers.
The three dimensions of the target pose are each discretized into 64 bins. $p(y|\textrm{ID})$, $p(z|\textrm{ID})$ and $p(\textrm{angle}|\textrm{ID})$ model the conditional distribution of a target pose given an object ID. The logits are also computed with linear transformations on features: $\textrm{logit}(y=j|\textrm{ID}=i) = g_{y}(h_t^i), j=1, 2, \cdots, 64, i=1, 2, \cdots, N$.
$V_{\phi}$ is a 2-layer MLP that computes a value scalar from the aggregated feature $\sum_{i=1}^{N+2} h_t^i$. 

The low-level controller is either an object teleporter that directly teleports the selected block to the target pose, or a sampling-based motion planner that generates collision-free trajectories of robot motions to fulfill the instructions given by the high-level designer.

\subsection{Phasic Policy Gradient}
Following~\cite{li2021learning}, our algorithm to train the RL designer is built upon Phasic Policy Gradient (PPG)~\cite{cobbe20ppg}. 
PPG alternates between the policy phase and the value phase. In the policy phase, we optimize $\psi$ and $\pi_{\theta}$ with the policy loss
\begin{equation*}
    \mathcal{L}^{\pi} = \mathbb{E}[\min(\rho \hat{A}, \textrm{clip}(\rho, 1-\epsilon, 1+\epsilon)\hat{A})], 
    \rho = \frac{\pi_{\theta}(a_t|\psi(s_t))}{\pi_{\theta_{old}}(a_t|\psi(s_t))},
\end{equation*}
where $\hat{A}$ is the advantage function. In the value phase, we train $V_{\phi}$ and $\psi$ by minimizing $\mathcal{L}^{joint}$ defined as 
\begin{equation*}
    \mathcal{L}^{joint} = \mathcal{L}^{V} + \beta_{clone} \mathcal{L}^{clone},  \mathcal{L}^V = \frac{1}{2}\mathbb{E}[(V_{\phi}(\psi(s_t)) - V^{targ})^2],
\end{equation*}
\begin{equation*}
    \mathcal{L}^{clone} = \mathbb{E}[KL[\pi_{\theta_{old}}(\cdot|\psi(s_t)) || \pi_{\theta}(\cdot|\psi(s_t))]].
\end{equation*}\label{sec:setup}

\section{Method}\label{sec:method}
Designing and constructing a bridge using varying-sized blocks is a challenging task due to sparse reward and complex physical constraints of stable bridges --- only a tiny subspace of possible object placements could yield stable architectures. 
Hence, we propose the {\name} technique, which adaptively resets the initial state to an intermediate configuration, to tackle this hard exploration challenge.
In addition, we improve the representation learning of the agent with a self-supervised auxiliary task, which can be shown to accelerate training significantly. Finally, we also fine-tune the pre-trained high-level designer with the integration of a low-level controller to get feasible instructions for the robot.

\begin{algorithm}[tb]
    \SetAlgoNoEnd
    \DontPrintSemicolon
    \SetAlgoLined
    \caption{PPG with prioritized memory reset}\label{algo:ptr}
    Intialize $\psi$, $\pi_{\theta}$, $V_{\phi}$. RL data buffer $\mathcal{B}$. An empty replay buffer $\mathcal{Q}$ to store reset states.\;
    $s_0 \sim \rho_0$\;
    \For{iter=0:n\_iters}{
        $\mathcal{B} \gets \emptyset$\;
        \For{t=0:n\_steps}{
            $\mathcal{Q}$.insert(priority($s_t$), $s_t$)\;
            $a_t \gets \pi_{\theta}(\psi(s_t))$\;
            $s_{t+1}, r_t, \textrm{terminate} \gets$ env.step($a_t$)\;
            $\mathcal{B} \gets \mathcal{B} \cup (s_t, a_t, r_t, s_{t+1}, \textrm{terminate})$\;
            \If{\textrm{terminate}}{
                \eIf{$rand() < p_{restart}$}{
                    $s_{t+1} \gets \mathcal{Q}$.pop()\;
                }{
                    $s_{t+1} \gets \rho_0$\;
                }
            } 
        }
        \For{batch\_data sampled from $\mathcal{B}$}{
            Optimize $\psi$ and $\pi_{\theta}$ with $\mathcal{L}^{\pi}$\;
        }
        \For{batch\_data sampled from $\mathcal{B}$}{
            Optimize $\psi$ and $V_{\phi}$ with $\mathcal{L}^{joint}$\;
        }
        Recompute priority($s$) for each $s$ in $\mathcal{Q}$\;
    }
\end{algorithm}
\subsection{Prioritized Memory Reset}
We propose {\name}, an adaptive reset paradigm to enhance the exploration ability of an RL agent. 
When an episode resets, we allow the agent to restart from an intermediate state it has previously visited with some probability instead of always from scratch. 
For on-policy RL methods, a training RL agent can occasionally reach some promising states that are close to success through random exploration. However, since each episode restarts from scratch in standard RL, the agent may not be able to re-visit those promising architectures. By contrast, {\name} enables the agent to directly teleport to these critical states without re-executing its policy.


The key design choice in {\name} is how to select critical states to restart from. In this work, we propose to use temporal difference (TD) error as a priority metric:
\begin{equation*}
    \textrm{priority}(s) = |r + \gamma V(s') - V(s)|,
\end{equation*}
where $s, r, s'$ are states, rewards, and the subsequent states collected from previous interactions with the environment, and $V$ is the learned value function.
Intuitively, the states that result in unexpected success or suddenly degrade to complete failure are with large TD errors. Restarting from these states could aggressively guide the agent to discover successful configurations and also practice to avoid catastrophic failure. 

We store the visited states and their priorities when interacting with the environment. When an episode terminates, we query the state with the highest priority and set it as the new initial state with probability $p_{restart}$; otherwise, we reset the episode with a random state sampled from the initial state distribution $\rho_0$. The framework of RL combined with {\name} is shown in Alg.~\ref{algo:ptr}.

\textbf{Implementation details: } We maintain all the visited states in a priority queue. 
Since our agent is constantly evolving, the stored priorities computed from old values would soon become stale. Therefore, we re-compute the priorities of all the tracked states after each training phase. We pop out the states with the least priorities when the priority queue is full.
\hide{
We store a diverse set of visited states, then sample the states from large temporal difference errors from this set as restart states. Intuitively, the agent may encounter some promising states of partially built structures during exploration although it does not reach a complete success. Allowing more trials from those states could lead to more reward signal, thus faster learning of meaningful behaviors. 
}
Since our state space is continuous and we can only restore a limited number of states, it is infeasible to keep track of all the visited states. Also, it is unnecessary to store multiple states that are similar to each other. Therefore, we adopt state hashing and only keep one representative for each hash value. The hash function we apply is defined as follows: we first get the heights of the built structure's upper surface, which we call the skyline vector, then discretize the vector as the hash key of the state.  
 
\begin{figure*}[tb]
    \centering
    \begin{minipage}{0.95\textwidth}
        \centering
        \includegraphics[width=\linewidth]{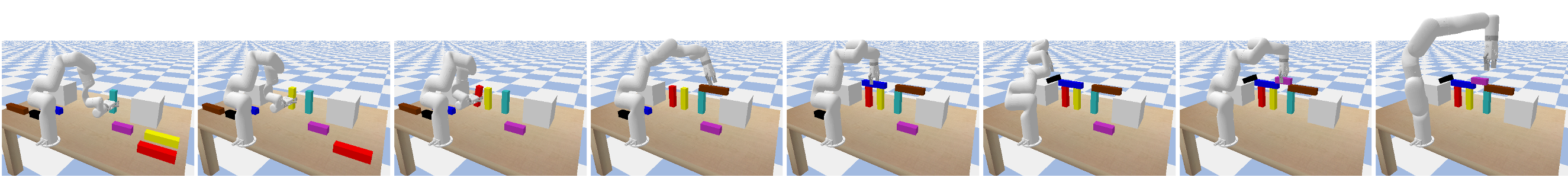}
        \includegraphics[width=\linewidth]{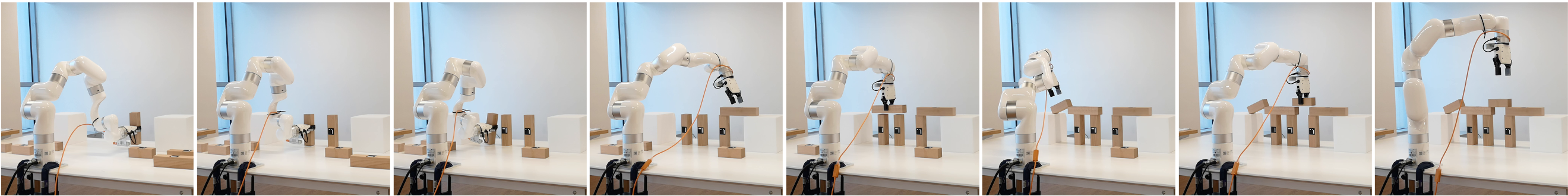}
    \end{minipage}
    \caption{Learned strategies for constructing a long bridge using 7 blocks with different sizes. The two rows are construction sequences in simulation and in the real world.}\label{fig:expr:policy_7b}
    \vspace{-5mm}
\end{figure*}

\subsection{Inverse Dynamics Prediction}
To better guide the training of the transformer-based encoder $\psi$, we propose to optimize a self-supervised auxiliary task jointly with the original RL objective. 
Given a transition tuple $(s_t, a_t, s_{t+1})$, the auxiliary task is to predict the action $a_t$ that results in the transition from $s_t$ to $s_{t+1}$, which is called inverse dynamics prediction. Any valid transitions from our environment can be used as training data for the auxiliary task. Therefore, the agent can always get rich supervision for learning representation even when the reward signal from the environment is very sparse. In our experiments, we optimize
\begin{equation*}
    \mathcal{L}^{joint} = \mathcal{L}^{V} + \beta_{clone}\mathcal{L}^{clone} + \beta_{aux}\mathcal{L}^{aux}
\end{equation*}
in the value phase of PPG training. $\mathcal{L}^{V}$ and $\mathcal{L}^{clone}$ are the original objectives in PPG, and $L^{aux}$ is the auxiliary prediction loss. 
We first compute the features $h_t^{[N+2]} = \psi(s_t^{[N+2]})$ and $h_{t+1}^{[N+2]} = \psi(s_{t+1}^{[N+2]})$ using the same policy feature extractor. Then we concatenate $h_t$ and $h_{t+1}$ together along the hidden size dimension to get $\hat{h}$. $\hat{h}$ is passed into an action predictor with almost the same architecture as $\pi_{\theta}$ to get $\hat{a}$ -- the difference is that the input linear layers become twice wider. Finally, $L^{aux}$ is computed as the cross-entropy between $\hat{a}$ and $a_t$.

Note that the designer only places a single block per step, so a large part of $s_{t+1}$ is often identical with $s_t$. Consequently, we choose inverse dynamics prediction as our self-supervised objective instead of another popular choice, forward prediction (i.e., predicting $s_{t+1}$ from $s_t$ and $a_t$), which may hurt representation learning and result in a degenerated policy. 


\subsection{Fine-tuning with Low-level Control Generator}
We first train the high-level designer with object teleportation, and then fine-tune the learned designer by integrating with a low-level motion generator. With an object teleporter, the environment directly repositions the selected block according to the instruction produced by the designer, runs forward simulation until all the objects become stable, and then takes the resulting stable state as the next state. This approach is originally proposed in \cite{li2021learning}. However, with random-sized objects, we empirically observe the learned designer often gives instructions infeasible for a robot arm to execute, e.g., it may instruct to pick up an object too close to another object for the robot arm to grasp without collision. Therefore, we further replace the object teleporter with a low-level motion generator, which fully simulates the movement of the robot arm, and fine-tune the pre-trained designer to generate feasible construction plans. 

We implement the low-level motion generator with a sampling-based motion planner. Each instruction from the designer is implemented as a pick-and-place task that grasps the block's center of mass. We divide each pick-and-place task into several sub-tasks: fetching the target object, changing the object pose, and moving back the arm. In each sub-task, we use a bidirectional RRT~\cite{kuffner2000rrt} motion planner to search a collision-free sequence of robot motions. If the planner fails to find a valid path (which may due to failing to find a grasp pose, target state in collision, etc.), the simulator will revert the scene to the state before the whole pick-and-place task and wait for the next instruction. 

Training the designer from scratch with integration of the low-level motion planner is much more time-consuming than with teleportation transition. We still try to train a designer from scratch with a mixed ratio of low-level controller and teleportation. Despite the long training time, this approach gets lower sample efficiency and fails to converge to a high success rate (more in Sec.~\ref{sec:expr:finetune}). We hypothesis that incorporating low-level control may hurt exploration in the early training stage of the designer in construction tasks. 


\subsection{Real Robot Deployment} 
We mount an xArm7 robot in the real world with the same configuration as in the simulated environment. A RealSense D435 RGBD camera is mounted on the hand of the robot. We attach ArUco markers~\cite{garrido2014automatic} to the building blocks to get accurate pose estimation. We estimate the lengths of the building blocks using contour approximation provided in OpenCV~\cite{opencv_library}. After parsing the scene at the beginning of an episode, the robot plans a sequence of joint angle positions with the trained high-level designer and the low-level motion generator, then executes along the planned trajectory.

\hide{
To narrow the gap between the simulation and the real world, we apply random force to the target object's center of mass when the robot releases that object. 
}

\section{Experiments}\label{sec:expr}
We first show how our method solves a bridge design and construction task with a collection of 7 random-sized blocks, then validate the effectiveness of all the algorithmic components with ablation studies. 
All the experiments in simulation are repeated over 3 seeds.
\begin{figure*}
    \centering
    \begin{minipage}{\textwidth}
        \begin{subfigure}[t]{0.43\textwidth}
            \centering
            \includegraphics[width=0.31\linewidth]{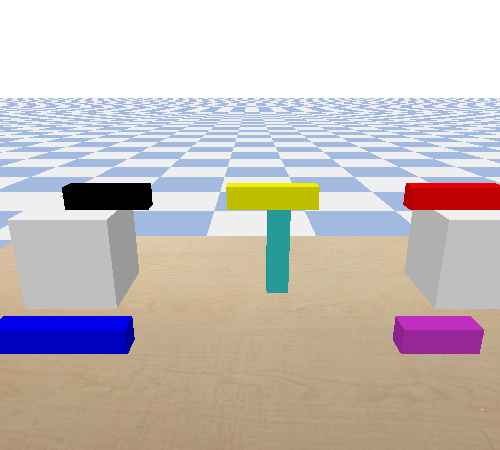}
            \includegraphics[width=0.31\linewidth]{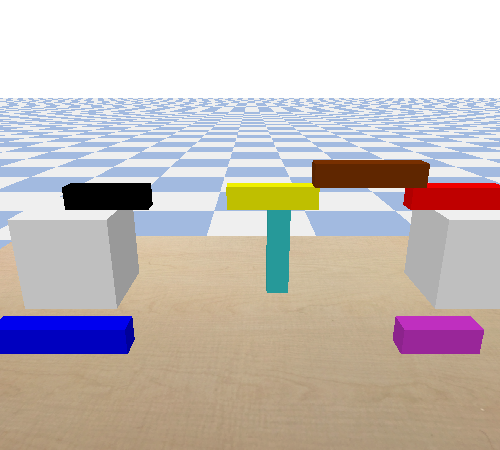}
            \includegraphics[width=0.31\linewidth]{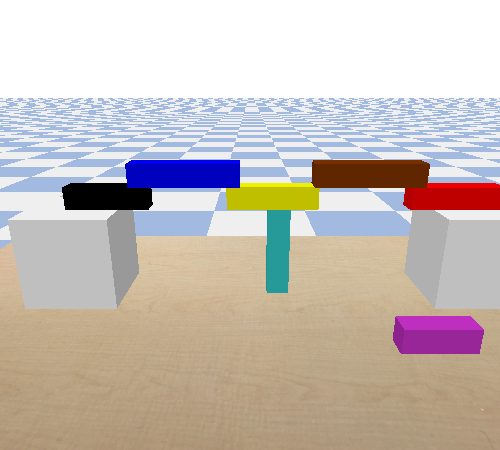}
            \caption{An efficient construction plan using 6 blocks.}
        \end{subfigure}
        \begin{subfigure}[t]{0.56\textwidth}
            \centering
            \includegraphics[width=0.24\linewidth]{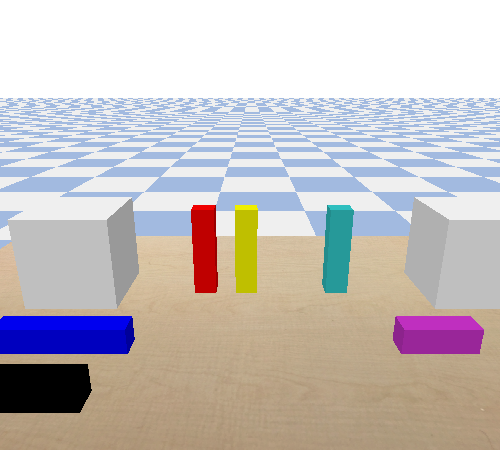}
            \includegraphics[width=0.24\linewidth]{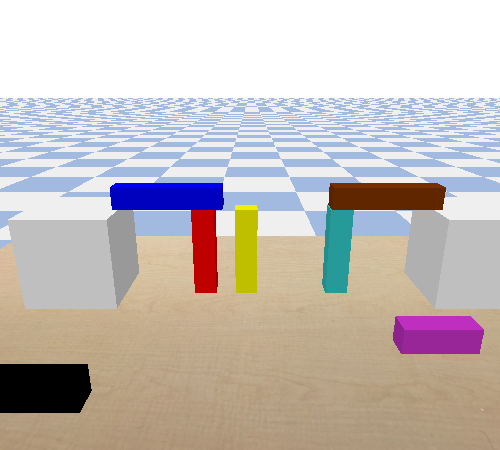}
            \includegraphics[width=0.24\linewidth]{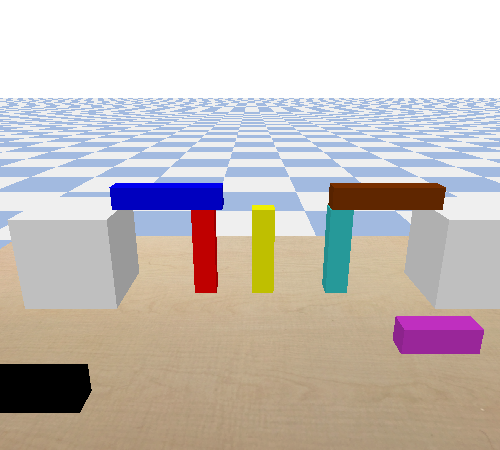}
            \includegraphics[width=0.24\linewidth]{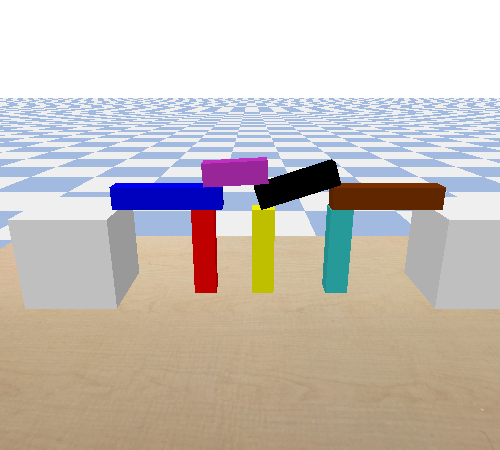}
            \caption{Another bridge design using all 7 blocks.}
        \end{subfigure}
    \end{minipage}
    \caption{Different modes of construction plans under the same task configuration.}
    \label{fig:expr:multimode}
    \vspace{-4mm}
\end{figure*}

\begin{figure*}
    \centering
    \includegraphics[width=0.13\textwidth]{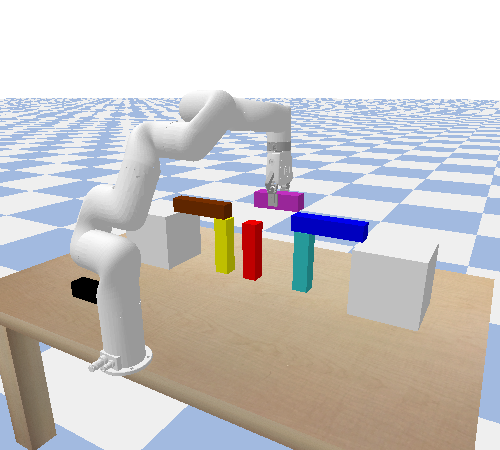}
    \includegraphics[width=0.13\textwidth]{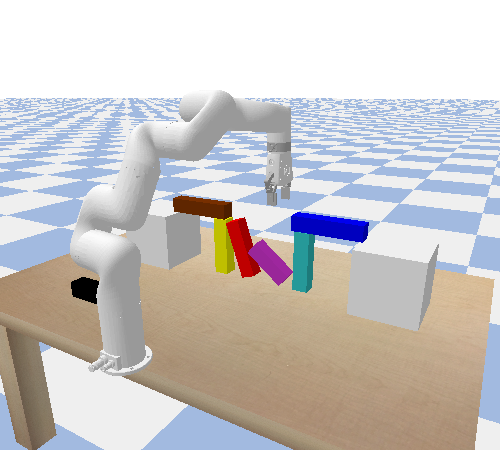}
    \includegraphics[width=0.13\textwidth]{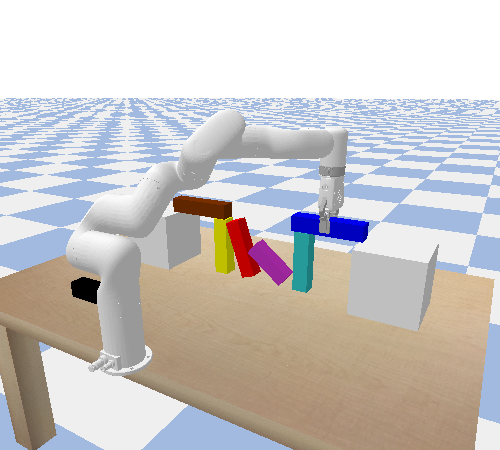}
    \includegraphics[width=0.13\textwidth]{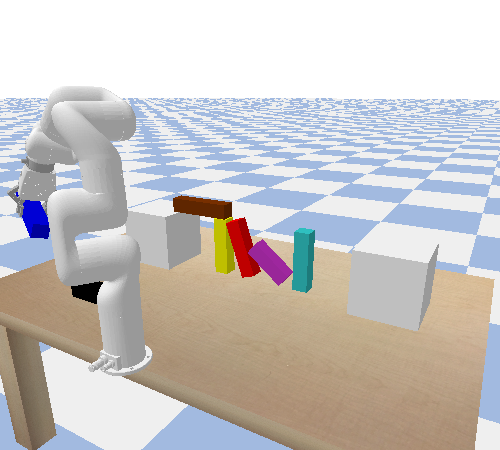}
    \includegraphics[width=0.13\textwidth]{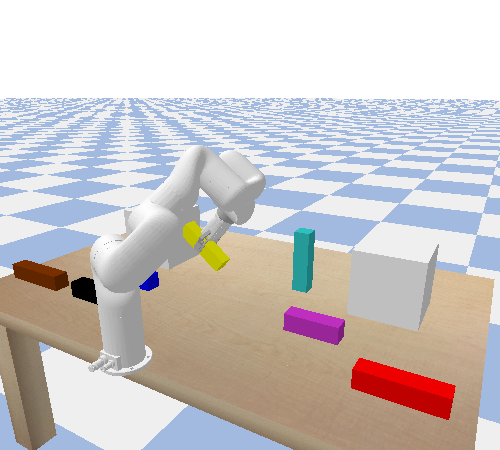}
    \includegraphics[width=0.13\textwidth]{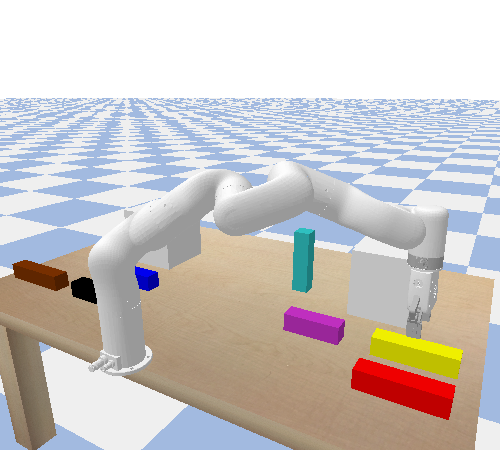}
    \includegraphics[width=0.13\textwidth]{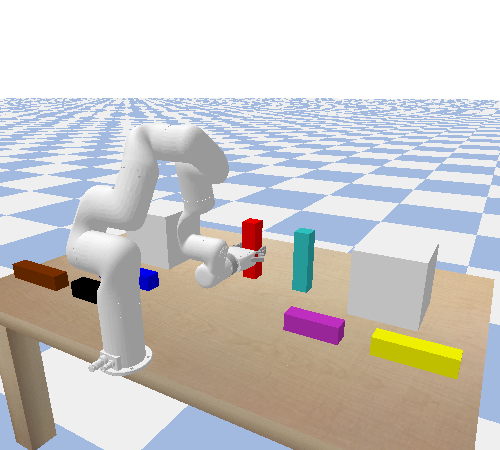}
    \caption{One failure case. The agent knocks down part of the built bridge when placing the short purple block, cleans up the messy scene by itself, and tries to build again. The trajectory terminates due to time limit.}
    \label{fig:expr:failure}
    \vspace{-2mm}
\end{figure*}


\subsection{Main Results}
We demonstrate how our agent designs and constructs a long bridge with a total of 7 building blocks in Fig.~\ref{fig:expr:policy_7b}. The building material set consists of three standard blocks with a length of 20cm, two short blocks of length 14cm, and two long blocks of length 24cm. The distance between cliffs is 65cm. The agent learns to put three standard blocks vertically inside the valley as supporting blocks, then put two long blocks on top of these supporting blocks as part of the bridge surface, finally fill in the gaps with two short blocks. 

\subsection{Visualization of Reset States with {\name}}
\begin{figure}[tb]
    \centering
    \begin{subfigure}[b]{0.3\linewidth}
        \centering
        \includegraphics[width=\linewidth]{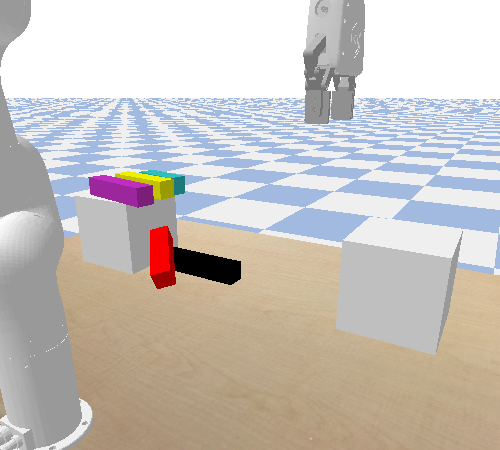}
        \caption{Iteration 10}
    \end{subfigure}
    \begin{subfigure}[b]{0.3\linewidth}
        \centering
        \includegraphics[width=\linewidth]{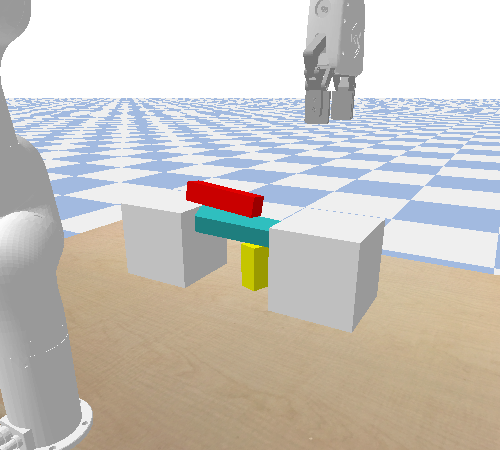}
        \caption{Iteration 50}
    \end{subfigure}
    \begin{subfigure}[b]{0.3\linewidth}
        \centering
        \includegraphics[width=\linewidth]{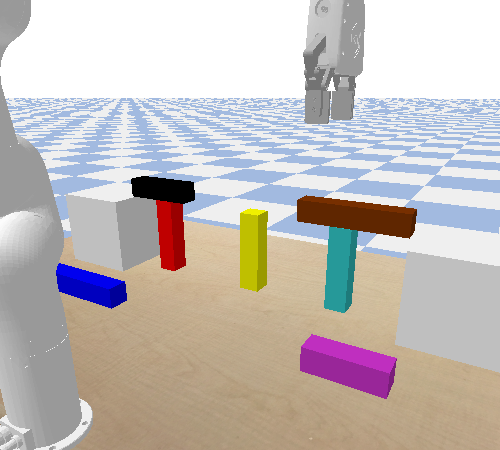}
        \caption{Iteration 100}
    \end{subfigure}
    \caption{
    Reset states with top priorities selected by {\name} at different training iterations.
    }
    \label{fig:reset_state}
    \vspace{-5mm}
\end{figure}

We visualize which states {\name} proposes to restart from. 
The states with the highest priorities during training are demonstrated in Fig.~\ref{fig:reset_state}. The quality of reset states is also evolving as the training proceeds. In the early stage, the states with large TD errors are messy states with objects randomly dropped in the scene. Then the agent gradually learns to construct more meaningful structures. Finally, the agent focuses on building very long bridges from partially built structures. The prioritized reset mechanism can be also viewed as an implicit curriculum for the agent. 

\subsection{Ablation Studies on High-level Designer Learning}
\begin{figure}
    \centering
    \includegraphics[width=0.9\linewidth]{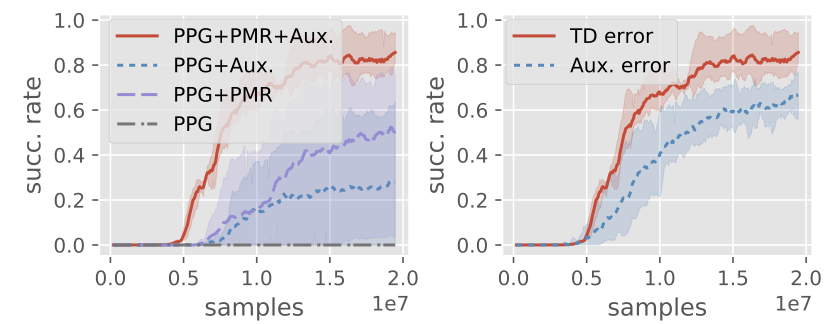}
    \caption{Ablation studies of different algorithm variants. The left figure shows the effectiveness of prioritized reset and auxiliary prediction task. Red: PPG with prioritized reset and auxiliary prediction task. Blue: PPG with auxiliary representation learning. Purple: PPG with prioritized reset. Grey: Pure PPG. The right figure compares the performances of different metrics to prioritized the reset states.}
    \label{fig:expr:ablation_blueprint}
    \vspace{-7mm}
\end{figure}

In this part, we directly use the object teleporter to execute the instructions given by the designer. 
All the experimented methods are evaluated on the tasks with a collection of 7 blocks of various sizes and are always reset from states sampled from $\rho_0$ in evaluation. 
The distance between cliffs is sampled from the range [2.75L, 3.75L].
We verify the effectiveness of {\name} and auxiliary task by comparing our method with the variants that remove one or both components. 
The ablation results are shown in the left half of Fig.~\ref{fig:expr:ablation_blueprint}.  
Our method (PPG + {\name} + auxiliary prediction, red) outperforms ``PPG + auxiliary prediction'' (blue) and ``PPG + {\name}'' (purple) with a large margin, indicating both {\name} and auxiliary prediction are critical for efficient learning of construction tasks. 
Note that naively apply PPG algorithm, which is essentially the method in~\cite{li2021learning}, leads to complete failure (grey). It also demonstrates this sparse-reward construction task with varying-sized building blocks is non-trivial for current on-policy RL algorithms to solve.

\begin{figure}
    \centering
    \includegraphics[width=0.6\linewidth]{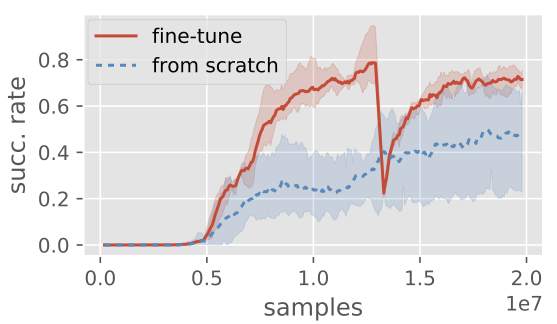}
    \caption{Comparison between pre-training high-level designer with teleportation then fine-tuning after integrated with a motion generator (red) and training from scratch combined with the low-level controller (blue). The first half of the red curve is evaluated with teleportation, and all other parts are evaluated with the motion generator. 
    }
    \label{fig:expr:finetune}
    \vspace{-5mm}
\end{figure}

We show performances of different metrics to prioritize the reset states in the right half of Fig.~\ref{fig:expr:ablation_blueprint}. The red curve is prioritizing with absolute TD error, and the blue curve is using the inverse dynamics prediction error to prioritize the states. Prioritizing with TD error achieves better sample efficiency and a higher success rate.

\subsection{Ablation Studies on Fine-tuning with Low-level Control}\label{sec:expr:finetune}
We first answer whether fine-tuning with a motion planner improves the performance of a pre-trained designer.
We take an RL designer trained with the object teleporter for $1.3e7$ timesteps, then continue training by combining it with the motion generator. 
As shown in Fig.~\ref{fig:expr:finetune}, directly executing the instructions of the pre-trained policy with the motion generator can only achieve a success rate of 0.224, but the success rate can be improved to 0.718 after fine-tuning. The blueprint policy can quickly adapt its instructions to avoid infeasible actions for the specific low-level controller, thus boost its performance especially at the beginning of the fine-tuning phase. 

We try another variant that trains the policy from scratch with mixed teleportation and motion generator. We use the motion generator to execute instructions with probability equal to the current success rate of the agent, and use the object teleporter otherwise. 
The overall sample efficiency is lower than pre-training then fine-tuning, and the success rate only converges to 0.472. The degradation of performance may due to insufficient exploration when integrated with the low-level controller in the early training stage. \hide{The performance gap between the agent integrated with low-level controller from scratch and the one pretrained with teleportation transition may due to lack of exploration when combined with the controller. }

\subsection{Learned Strategies and Failure Cases}
Our agent can discover multiple solutions for the same task configuration. We set the distance between cliffs to be 69cm, and give the agent 7 blocks of length 14cm, 18cm, 20cm, 20cm, 20cm, 24cm, 24cm. In Fig.~\ref{fig:expr:multimode}, the agent discovers two different construction plans. The first solution only uses 6 blocks to solve the task. In the second solution, the agent utilizes all 7 blocks. It strategically adjusts the position of the yellow block before putting the short black block.

A failure case of the agent is depicted in Fig.~\ref{fig:expr:failure}. The agent intends to connect the red and light blue blocks with the short purple block, but knocks down other blocks when dropping the pink block from the air. The agent then spends many steps to clear the scene, and tries to construct again. The agent fails to reach a successful state before the episode terminates.

\section{Conclusion} \label{sec:conclusion}
We tackle a challenging sparse-reward manipulation task that designs and constructs bridges with varying-sized building blocks. We propose a novel learning paradigm {\name}, that allows the agent to restart from critical states it has visited before to deal with the exploration issue. We additionally propose an auxiliary representation learning task and fine-tuning with integration of a motion generator to successfully build a system that can construct interesting structures using building blocks of various sizes in the real world. 


\bibliographystyle{IEEEtran}
\bibliography{references}

\begin{thebibliography}{10}
\providecommand{\url}[1]{#1}
\csname url@samestyle\endcsname
\providecommand{\newblock}{\relax}
\providecommand{\bibinfo}[2]{#2}
\providecommand{\BIBentrySTDinterwordspacing}{\spaceskip=0pt\relax}
\providecommand{\BIBentryALTinterwordstretchfactor}{4}
\providecommand{\BIBentryALTinterwordspacing}{\spaceskip=\fontdimen2\font plus
\BIBentryALTinterwordstretchfactor\fontdimen3\font minus
  \fontdimen4\font\relax}
\providecommand{\BIBforeignlanguage}[2]{{%
\expandafter\ifx\csname l@#1\endcsname\relax
\typeout{** WARNING: IEEEtran.bst: No hyphenation pattern has been}%
\typeout{** loaded for the language `#1'. Using the pattern for}%
\typeout{** the default language instead.}%
\else
\language=\csname l@#1\endcsname
\fi
#2}}
\providecommand{\BIBdecl}{\relax}
\BIBdecl

\bibitem{gu2017deep}
S.~Gu, E.~Holly, T.~Lillicrap, and S.~Levine, ``Deep reinforcement learning for
  robotic manipulation with asynchronous off-policy updates,'' in \emph{2017
  IEEE international conference on robotics and automation (ICRA)}.\hskip 1em
  plus 0.5em minus 0.4em\relax IEEE, 2017, pp. 3389--3396.

\bibitem{rajeswaran18dexterous}
A.~Rajeswaran, V.~Kumar, A.~Gupta, G.~Vezzani, J.~Schulman, E.~Todorov, and
  S.~Levine, ``Learning complex dexterous manipulation with deep reinforcement
  learning and demonstrations,'' in \emph{Robotics: Science and Systems XIV,
  Carnegie Mellon University, Pittsburgh, Pennsylvania, USA, June 26-30, 2018},
  H.~Kress{-}Gazit, S.~S. Srinivasa, T.~Howard, and N.~Atanasov, Eds., 2018.

\bibitem{akkaya2019solving}
I.~Akkaya, M.~Andrychowicz, M.~Chociej, M.~Litwin, B.~McGrew, A.~Petron,
  A.~Paino, M.~Plappert, G.~Powell, R.~Ribas \emph{et~al.}, ``Solving rubik's
  cube with a robot hand,'' \emph{arXiv preprint arXiv:1910.07113}, 2019.

\bibitem{yu2020meta}
T.~Yu, D.~Quillen, Z.~He, R.~Julian, K.~Hausman, C.~Finn, and S.~Levine,
  ``Meta-world: A benchmark and evaluation for multi-task and meta
  reinforcement learning,'' in \emph{Conference on Robot Learning}.\hskip 1em
  plus 0.5em minus 0.4em\relax PMLR, 2020, pp. 1094--1100.

\bibitem{kalashnikov2018qt}
D.~Kalashnikov, A.~Irpan, P.~Pastor, J.~Ibarz, A.~Herzog, E.~Jang, D.~Quillen,
  E.~Holly, M.~Kalakrishnan, V.~Vanhoucke \emph{et~al.}, ``Qt-opt: Scalable
  deep reinforcement learning for vision-based robotic manipulation,''
  \emph{arXiv preprint arXiv:1806.10293}, 2018.

\bibitem{nair18overcoming}
A.~Nair, B.~McGrew, M.~Andrychowicz, W.~Zaremba, and P.~Abbeel, ``Overcoming
  exploration in reinforcement learning with demonstrations,'' in \emph{2018
  {IEEE} International Conference on Robotics and Automation, {ICRA} 2018,
  Brisbane, Australia, May 21-25, 2018}.\hskip 1em plus 0.5em minus 0.4em\relax
  {IEEE}, 2018, pp. 6292--6299.

\bibitem{li2020towards}
R.~Li, A.~Jabri, T.~Darrell, and P.~Agrawal, ``Towards practical multi-object
  manipulation using relational reinforcement learning,'' in \emph{2020 IEEE
  International Conference on Robotics and Automation (ICRA)}.\hskip 1em plus
  0.5em minus 0.4em\relax IEEE, 2020, pp. 4051--4058.

\bibitem{openai2021asymmetric}
O.~OpenAI, M.~Plappert, R.~Sampedro, T.~Xu, I.~Akkaya, V.~Kosaraju,
  P.~Welinder, R.~D'Sa, A.~Petron, H.~P. d.~O. Pinto \emph{et~al.},
  ``Asymmetric self-play for automatic goal discovery in robotic
  manipulation,'' \emph{arXiv preprint arXiv:2101.04882}, 2021.

\bibitem{li2019hrl4in}
C.~Li, F.~Xia, R.~Mart{\'{\i}}n{-}Mart{\'{\i}}n, and S.~Savarese, ``{HRL4IN:}
  hierarchical reinforcement learning for interactive navigation with mobile
  manipulators,'' in \emph{3rd Annual Conference on Robot Learning, CoRL 2019,
  Osaka, Japan, October 30 - November 1, 2019, Proceedings}, ser. Proceedings
  of Machine Learning Research, L.~P. Kaelbling, D.~Kragic, and K.~Sugiura,
  Eds., vol. 100.\hskip 1em plus 0.5em minus 0.4em\relax {PMLR}, 2019, pp.
  603--616.

\bibitem{xiarelmogen}
F.~Xia, C.~Li, R.~Mart{\i}n-Mart{\i}n, O.~Litany, A.~Toshev, and S.~Savarese,
  ``Relmogen: Integrating motion generation in reinforcement learning for
  mobile manipulation.''

\bibitem{balaguer2011combining}
B.~Balaguer and S.~Carpin, ``Combining imitation and reinforcement learning to
  fold deformable planar objects,'' in \emph{2011 {IEEE/RSJ} International
  Conference on Intelligent Robots and Systems, {IROS} 2011, San Francisco, CA,
  USA, September 25-30, 2011}.\hskip 1em plus 0.5em minus 0.4em\relax {IEEE},
  2011, pp. 1405--1412.

\bibitem{luo19assembly}
J.~{Luo}, E.~{Solowjow}, C.~{Wen}, J.~A. {Ojea}, A.~M. {Agogino}, A.~{Tamar},
  and P.~{Abbeel}, ``Reinforcement learning on variable impedance controller
  for high-precision robotic assembly,'' in \emph{2019 International Conference
  on Robotics and Automation (ICRA)}, 2019, pp. 3080--3087.

\bibitem{zhu2020hierarchical}
Y.~Zhu, J.~Tremblay, S.~Birchfield, and Y.~Zhu, ``Hierarchical planning for
  long-horizon manipulation with geometric and symbolic scene graphs,''
  \emph{arXiv preprint arXiv:2012.07277}, 2020.

\bibitem{li2021learning}
Y.~Li, T.~Kong, L.~Li, Y.~Li, and Y.~Wu, ``Learning to design and construct
  bridge without blueprint,'' in \emph{2021 IEEE/RSJ International Conference
  on Intelligent Robots and Systems (IROS)}.\hskip 1em plus 0.5em minus
  0.4em\relax IEEE, 2021, pp. 2398--2405.

\bibitem{florensa2017reverse}
C.~Florensa, D.~Held, M.~Wulfmeier, M.~Zhang, and P.~Abbeel, ``Reverse
  curriculum generation for reinforcement learning,'' in \emph{Conference on
  robot learning}.\hskip 1em plus 0.5em minus 0.4em\relax PMLR, 2017, pp.
  482--495.

\bibitem{florensa2018automatic}
C.~Florensa, D.~Held, X.~Geng, and P.~Abbeel, ``Automatic goal generation for
  reinforcement learning agents,'' in \emph{International Conference on Machine
  Learning}, 2018, pp. 1515--1528.

\bibitem{knepper2013ikeabot}
R.~A. Knepper, T.~Layton, J.~Romanishin, and D.~Rus, ``Ikeabot: An autonomous
  multi-robot coordinated furniture assembly system,'' in \emph{2013 IEEE
  International conference on robotics and automation}.\hskip 1em plus 0.5em
  minus 0.4em\relax IEEE, 2013, pp. 855--862.

\bibitem{nagele20lego}
L.~{Nägele}, A.~{Hoffmann}, A.~{Schierl}, and W.~{Reif}, ``Legobot: Automated
  planning for coordinated multi-robot assembly of lego structures*,'' in
  \emph{2020 IEEE/RSJ International Conference on Intelligent Robots and
  Systems (IROS)}, 2020, pp. 9088--9095.

\bibitem{zakka2020form2fit}
K.~Zakka, A.~Zeng, J.~Lee, and S.~Song, ``Form2fit: Learning shape priors for
  generalizable assembly from disassembly,'' in \emph{2020 IEEE International
  Conference on Robotics and Automation (ICRA)}.\hskip 1em plus 0.5em minus
  0.4em\relax IEEE, 2020, pp. 9404--9410.

\bibitem{haarnoja2018composable}
T.~Haarnoja, V.~Pong, A.~Zhou, M.~Dalal, P.~Abbeel, and S.~Levine, ``Composable
  deep reinforcement learning for robotic manipulation,'' in \emph{2018 IEEE
  International Conference on Robotics and Automation (ICRA)}.\hskip 1em plus
  0.5em minus 0.4em\relax IEEE, 2018, pp. 6244--6251.

\bibitem{batra20rearrangement}
\BIBentryALTinterwordspacing
D.~Batra, A.~X. Chang, S.~Chernova, A.~J. Davison, J.~Deng, V.~Koltun,
  S.~Levine, J.~Malik, I.~Mordatch, R.~Mottaghi, M.~Savva, and H.~Su,
  ``Rearrangement: {A} challenge for embodied {AI},'' \emph{CoRR}, vol.
  abs/2011.01975, 2020. [Online]. Available:
  \url{https://arxiv.org/abs/2011.01975}
\BIBentrySTDinterwordspacing

\bibitem{ritchie2015generating}
D.~Ritchie, S.~Lin, N.~D. Goodman, and P.~Hanrahan, ``Generating design
  suggestions under tight constraints with gradient-based probabilistic
  programming,'' in \emph{Computer Graphics Forum}, vol.~34, no.~2.\hskip 1em
  plus 0.5em minus 0.4em\relax Wiley Online Library, 2015, pp. 515--526.

\bibitem{ritchie2015controlling}
D.~Ritchie, B.~Mildenhall, N.~D. Goodman, and P.~Hanrahan, ``Controlling
  procedural modeling programs with stochastically-ordered sequential monte
  carlo,'' \emph{ACM Transactions on Graphics (TOG)}, vol.~34, no.~4, pp.
  1--11, 2015.

\bibitem{kulkarni2016hierarchical}
T.~D. Kulkarni, K.~Narasimhan, A.~Saeedi, and J.~Tenenbaum, ``Hierarchical deep
  reinforcement learning: Integrating temporal abstraction and intrinsic
  motivation,'' \emph{Advances in neural information processing systems},
  vol.~29, pp. 3675--3683, 2016.

\bibitem{bacon17option}
\BIBentryALTinterwordspacing
P.~Bacon, J.~Harb, and D.~Precup, ``The option-critic architecture,'' in
  \emph{Proceedings of the Thirty-First {AAAI} Conference on Artificial
  Intelligence, February 4-9, 2017, San Francisco, California, {USA}}, S.~P.
  Singh and S.~Markovitch, Eds.\hskip 1em plus 0.5em minus 0.4em\relax {AAAI}
  Press, 2017, pp. 1726--1734. [Online]. Available:
  \url{http://aaai.org/ocs/index.php/AAAI/AAAI17/paper/view/14858}
\BIBentrySTDinterwordspacing

\bibitem{nachum18data-efficient}
O.~Nachum, S.~Gu, H.~Lee, and S.~Levine, ``Data-efficient hierarchical
  reinforcement learning,'' in \emph{Advances in Neural Information Processing
  Systems 31: Annual Conference on Neural Information Processing Systems 2018,
  NeurIPS 2018, December 3-8, 2018, Montr{\'{e}}al, Canada}, S.~Bengio, H.~M.
  Wallach, H.~Larochelle, K.~Grauman, N.~Cesa{-}Bianchi, and R.~Garnett, Eds.,
  2018, pp. 3307--3317.

\bibitem{nachum19near}
\BIBentryALTinterwordspacing
------, ``Near-optimal representation learning for hierarchical reinforcement
  learning,'' in \emph{7th International Conference on Learning
  Representations, {ICLR} 2019, New Orleans, LA, USA, May 6-9, 2019}.\hskip 1em
  plus 0.5em minus 0.4em\relax OpenReview.net, 2019. [Online]. Available:
  \url{https://openreview.net/forum?id=H1emus0qF7}
\BIBentrySTDinterwordspacing

\bibitem{bagaria2020option}
\BIBentryALTinterwordspacing
A.~Bagaria and G.~Konidaris, ``Option discovery using deep skill chaining,'' in
  \emph{International Conference on Learning Representations}, 2020. [Online].
  Available: \url{https://openreview.net/forum?id=B1gqipNYwH}
\BIBentrySTDinterwordspacing

\bibitem{heess16learning}
\BIBentryALTinterwordspacing
N.~Heess, G.~Wayne, Y.~Tassa, T.~P. Lillicrap, M.~A. Riedmiller, and D.~Silver,
  ``Learning and transfer of modulated locomotor controllers,'' \emph{CoRR},
  vol. abs/1610.05182, 2016. [Online]. Available:
  \url{http://arxiv.org/abs/1610.05182}
\BIBentrySTDinterwordspacing

\bibitem{zhang2020automatic}
Y.~Zhang, P.~Abbeel, and L.~Pinto, ``Automatic curriculum learning through
  value disagreement,'' \emph{Advances in Neural Information Processing
  Systems}, vol.~33, 2020.

\bibitem{sukhbaatar2018intrinsic}
S.~Sukhbaatar, Z.~Lin, I.~Kostrikov, G.~Synnaeve, A.~Szlam, and R.~Fergus,
  ``Intrinsic motivation and automatic curricula via asymmetric self-play,'' in
  \emph{International Conference on Learning Representations}, 2018.

\bibitem{singh2005intrinsically}
S.~Singh, A.~G. Barto, and N.~Chentanez, ``Intrinsically motivated
  reinforcement learning,'' MASSACHUSETTS UNIV AMHERST DEPT OF COMPUTER
  SCIENCE, Tech. Rep., 2005.

\bibitem{burda19rnd}
\BIBentryALTinterwordspacing
Y.~Burda, H.~Edwards, A.~J. Storkey, and O.~Klimov, ``Exploration by random
  network distillation,'' in \emph{7th International Conference on Learning
  Representations, {ICLR} 2019, New Orleans, LA, USA, May 6-9, 2019}.\hskip 1em
  plus 0.5em minus 0.4em\relax OpenReview.net, 2019. [Online]. Available:
  \url{https://openreview.net/forum?id=H1lJJnR5Ym}
\BIBentrySTDinterwordspacing

\bibitem{pathak18largescale}
Y.~Burda, H.~Edwards, D.~Pathak, A.~Storkey, T.~Darrell, and A.~A. Efros,
  ``Large-scale study of curiosity-driven learning,'' in \emph{ICLR}, 2019.

\bibitem{ecoffet2021first}
A.~Ecoffet, J.~Huizinga, J.~Lehman, K.~O. Stanley, and J.~Clune, ``First
  return, then explore,'' \emph{Nature}, vol. 590, no. 7847, pp. 580--586,
  2021.

\bibitem{schrittwieser2020mastering}
J.~Schrittwieser, I.~Antonoglou, T.~Hubert, K.~Simonyan, L.~Sifre, S.~Schmitt,
  A.~Guez, E.~Lockhart, D.~Hassabis, T.~Graepel \emph{et~al.}, ``Mastering
  atari, go, chess and shogi by planning with a learned model,'' \emph{Nature},
  vol. 588, no. 7839, pp. 604--609, 2020.

\bibitem{pathak17curiosity}
D.~Pathak, P.~Agrawal, A.~A. Efros, and T.~Darrell, ``Curiosity-driven
  exploration by self-supervised prediction,'' in \emph{Proceedings of the 34th
  International Conference on Machine Learning, {ICML} 2017, Sydney, NSW,
  Australia, 6-11 August 2017}, ser. Proceedings of Machine Learning Research,
  D.~Precup and Y.~W. Teh, Eds., vol.~70.\hskip 1em plus 0.5em minus
  0.4em\relax {PMLR}, 2017, pp. 2778--2787.

\bibitem{hansen21self}
N.~Hansen, R.~Jangir, Y.~Sun, G.~Aleny{\`{a}}, P.~Abbeel, A.~A. Efros,
  L.~Pinto, and X.~Wang, ``Self-supervised policy adaptation during
  deployment,'' in \emph{9th International Conference on Learning
  Representations, {ICLR} 2021, Virtual Event, Austria, May 3-7, 2021}.\hskip
  1em plus 0.5em minus 0.4em\relax OpenReview.net, 2021.

\bibitem{coumans2020}
E.~Coumans and Y.~Bai, ``Pybullet, a python module for physics simulation for
  games, robotics and machine learning,'' \url{http://pybullet.org},
  2016--2020.

\bibitem{Vaswani17attention}
A.~Vaswani, N.~Shazeer, N.~Parmar, J.~Uszkoreit, L.~Jones, A.~N. Gomez,
  L.~Kaiser, and I.~Polosukhin, ``Attention is all you need,'' in
  \emph{Advances in Neural Information Processing Systems 30: Annual Conference
  on Neural Information Processing Systems 2017, December 4-9, 2017, Long
  Beach, CA, {USA}}, I.~Guyon, U.~von Luxburg, S.~Bengio, H.~M. Wallach,
  R.~Fergus, S.~V.~N. Vishwanathan, and R.~Garnett, Eds., 2017, pp. 5998--6008.

\bibitem{cobbe20ppg}
K.~Cobbe, J.~Hilton, O.~Klimov, and J.~Schulman, ``Phasic policy gradient,''
  \emph{CoRR}, vol. abs/2009.04416, 2020.

\bibitem{kuffner2000rrt}
J.~J. Kuffner and S.~M. LaValle, ``Rrt-connect: An efficient approach to
  single-query path planning,'' in \emph{Proceedings 2000 ICRA. Millennium
  Conference. IEEE International Conference on Robotics and Automation.
  Symposia Proceedings (Cat. No. 00CH37065)}, vol.~2.\hskip 1em plus 0.5em
  minus 0.4em\relax IEEE, 2000, pp. 995--1001.

\bibitem{garrido2014automatic}
S.~Garrido-Jurado, R.~Mu{\~n}oz-Salinas, F.~J. Madrid-Cuevas, and M.~J.
  Mar{\'\i}n-Jim{\'e}nez, ``Automatic generation and detection of highly
  reliable fiducial markers under occlusion,'' \emph{Pattern Recognition},
  vol.~47, no.~6, pp. 2280--2292, 2014.

\bibitem{opencv_library}
G.~Bradski, ``{The OpenCV Library},'' \emph{Dr. Dobb's Journal of Software
  Tools}, 2000.

\end{thebibliography}

\end{document}